\documentclass[10pt,twocolumn,letterpaper]{article}

\usepackage[pagenumbers]{cvpr} 

\usepackage{graphicx}
\usepackage{amsmath}
\usepackage{amssymb}
\usepackage{booktabs}

\usepackage[pagebackref,breaklinks,colorlinks]{hyperref}

\usepackage[capitalize]{cleveref}
\crefname{section}{Sec.}{Secs.}
\Crefname{section}{Section}{Sections}
\Crefname{table}{Table}{Tables}
\crefname{table}{Tab.}{Tabs.}

\def\cvprPaperID{*****} 
\def\confName{CVPR}
\def\confYear{2022}

\begin{document}

\title{Generalizable Method for Face Anti-Spoofing with Semi-Supervised Learning}

\author{Nikolay Sergievskiy\\
XIX.ai\\
{\tt\small nikolay@getentry.com}
\and
Roman Vlasov\\
XIX.ai\\
{\tt\small romanvlasov@getentry.com}
\and
Roman Trusov\\
XIX.ai\\
{\tt\small roman@getentry.com}
}
\maketitle

\begin{abstract}

    Face anti-spoofing has drawn a lot of attention due to the high security requirements in biometric authentication systems. Bringing face biometric to commercial hardware became mostly dependent on developing reliable methods for detecting fake login sessions without specialized sensors. Current CNN-based method perform well on the domains they were trained for, but often show poor generalization on previously unseen datasets. In this paper we describe a method for utilizing unsupervised pretraining for improving performance across multiple datasets without any adaptation, introduce the Entry Antispoofing Dataset for supervised fine-tuning, and propose a multi-class auxiliary classification layer for augmenting the binary classification task of detecting spoofing attempts with explicit interpretable signals. We demonstrate the efficiency of our model by achieving state-of-the-art results on cross-dataset testing on MSU-MFSD, Replay-Attack, and OULU-NPU datasets.
\end{abstract}

\section{Introduction}
\label{sec:intro}
Biometric authentication systems based on face recognition are taking prominence in both everyday life and high-value transactions with strong security requirements. However, despite the recent advances in computer vision, these systems are still confined to specialized hardware that utilizes depth or NIR sensors for preventing Presentation Attacks. Among the Presentation Attacks, print and replay attacks are the most common, and detecting them became an important problem in the field of biometric authentication. From the practical application perspective, developing a robust algorithm for detecting such attacks on commercial webcams using only signals from the video stream would enable wide adoption of face-based authentication and verification. 

Recently published results show that deep learning-based models can achieve good results on the datasets they were trained on \cite{CDCNN}\cite{CNN-LSTM}\cite{3d-synth}, but the generalization to other datasets is not so easily achieved - i.e. when the model that achieved a state-of-the-art result on benchmark A is tested on benchmark B (the protocol we will be referring to as "cross-test"), the discrepancy in scores is quite significant, even with the latest breakthroughs in domain adaptation that were aimed to address this problem. The consequences of such discrepancy in the real world are quite damaging for the application security.

The key motivation behind this work is that achieving strong generalization on cross-testing on multiple string benchmarks would reliably reflect the effectiveness of the algorithm in the wild. We propose to achieve this generalization by changing the approach to collecting training data. Moreover, motivated by the previous work in self-supervised learning\cite{big-self-supervised}\cite{SEER}, we experiment with the network pretraining on larger datasets to improve the results further.

To validate the effectiveness of the developed method, we report the evaluation results in two experimental settings: intra-dataset test is evaluated on a test portion of our internal dataset, and cross-test is evaluated on well-known and established benchmarks: MSU-MFSD\cite{MFSD}, Replay-Attack\cite{mci/Chingovska2012}, and OULU-NPU\cite{OULU_NPU_2017}.

\subsection{Contribution}
In this paper we present the following results:

\begin{enumerate}
    \item Large-scale dataset with additional semantic annotation for training anti-spoofing models, together with an efficient approach to labeled data collection.
    \item Effectiveness of task-agnostic unsupervised pretraining and explicitly defined spoofing attributes as a part of training objective.
    \item New state of the art results on MSU-MFSD\cite{MFSD} and Replay-Attack\cite{mci/Chingovska2012} cross-dataset tests (with the results surpassing even the best \textit{intra-tests}), and cross-test on OULU-NPU\cite{OULU_NPU_2017}. Most importantly, we demonstrate consistently strong results across the most difficult face anti-spoofing benchmarks, which indicates low degree of domain overfitting.
\end{enumerate}

\section{Related work}

Recent publications on task-agnostic self-supervised and semi-supervised pretraining show that using large unlabeled datasets for unsupervised pretraining followed by supervised fine-tuning is capable of outperforming standard supervised learning methods \cite{big-self-supervised} \cite{SEER}. This is especially promising for the field of face anti-spoofing, where the problem of lack of comprehensive labeled datasets suitable for building models viable for security applications is especially severe. There have been efforts to alleviate this problem with using rich semantic annotations \cite{CelebA-Spoof}. Alternative paradigms of circumventing the data shortage by domain adaptation and generating synthetic data were demonstrated by \cite{GFA-CNN} \cite{3d-synth}, citing the problem of domain shift as one of the most critical for anti-spoofing. 

The main problem of existing methods is still in the domain shift and the lack of generalization between different datasets, as shown in cross-dataset tests, even in works that demonstrate state-of-the-art results on intra-dataset tests \cite{GFA-CNN} \cite{3d-synth} \cite{CNN-LSTM} \cite{CDCNN}. 

\section{Method}

\subsection{Entry Antispoofing Dataset}

Existing datasets for face anti-spoofing cover too narrow a domain, compared to the diversity of camera/lighting/distance/attack conditions seen in the real world. While achieving low error scores on open-source benchmarks using a cross-dataset protocol is indicative of good performance of a network under some subset of conditions, it turned out to be an unreliable predictor of the stability and accuracy when used in a real application. Specifically, regardless of the Attack Presentation Classification Error Rate (APCER) and Bona Fide Presentation Classification Error Rate (BPCER) shown by a candidate model trained on open-source datasets, there always were multiple sets of conditions where the model's predictions started being inconsistent. We have addressed this problem by building an internal dataset that would consist of the training portion and a test subset that would be comprehensive enough to address the missing subdomains in other benchmarks.

Entry Antispoofing Dataset consists of 83000 live video recordings collected via a custom-built UI that simulates the process of logging into a web-based biometric authentication system like Entry. The recordings were collected and labeled via a crowdsourcing data labeling service, from 45000 participants from more than 20 countries on five continents. 
Each recording was made on a mobile or laptop webcam (with roughly $30\%$ of recordings being from laptop cameras, and $70\%$ - from mobile), with subject's face visible from multiple angles. Subjects' genders, ages and ethnicities are not correlated with their collected recordings being spoofing or bona fide sessions, but overall distribution was not restricted in order to be as close to real-life demographic of potential users as possible. The process of collecting the dataset was iterative, with each version of the dataset produced after identifying a "blind spot subdomain" — specific set of camera/lighting/distance/other conditions, that were leading to unstable performance of the model. These blind spots were identified by crowdsourcing attacks on different iterations of earlier anti-spoofing models that were provided by our research team.  

\begin{table}[h]
  \centering
  \begin{tabular}{l | c | c}
    \toprule
    \textbf{Dataset} & \textbf{Videos} & \textbf{Subjects} \\
    \midrule
    \midrule
    \textbf{Entry} & \textbf{83000} & \textbf{45000} \\
    MSU-MFSD\cite{MFSD}      & 280  & 35 \\
    Replay-Attack\cite{mci/Chingovska2012} & 1300 & 50 \\
    OULU-NPU\cite{OULU_NPU_2017}      & 4950 & 55 \\
    \bottomrule
  \end{tabular}
  \caption{Comparison of existing datasets for face anti-spoofing.}
  \label{datasets}
\end{table}

\subsection{Architecture and algorithm}

\begin{figure}[h]
  \centering
  \includegraphics[width=0.9\linewidth]{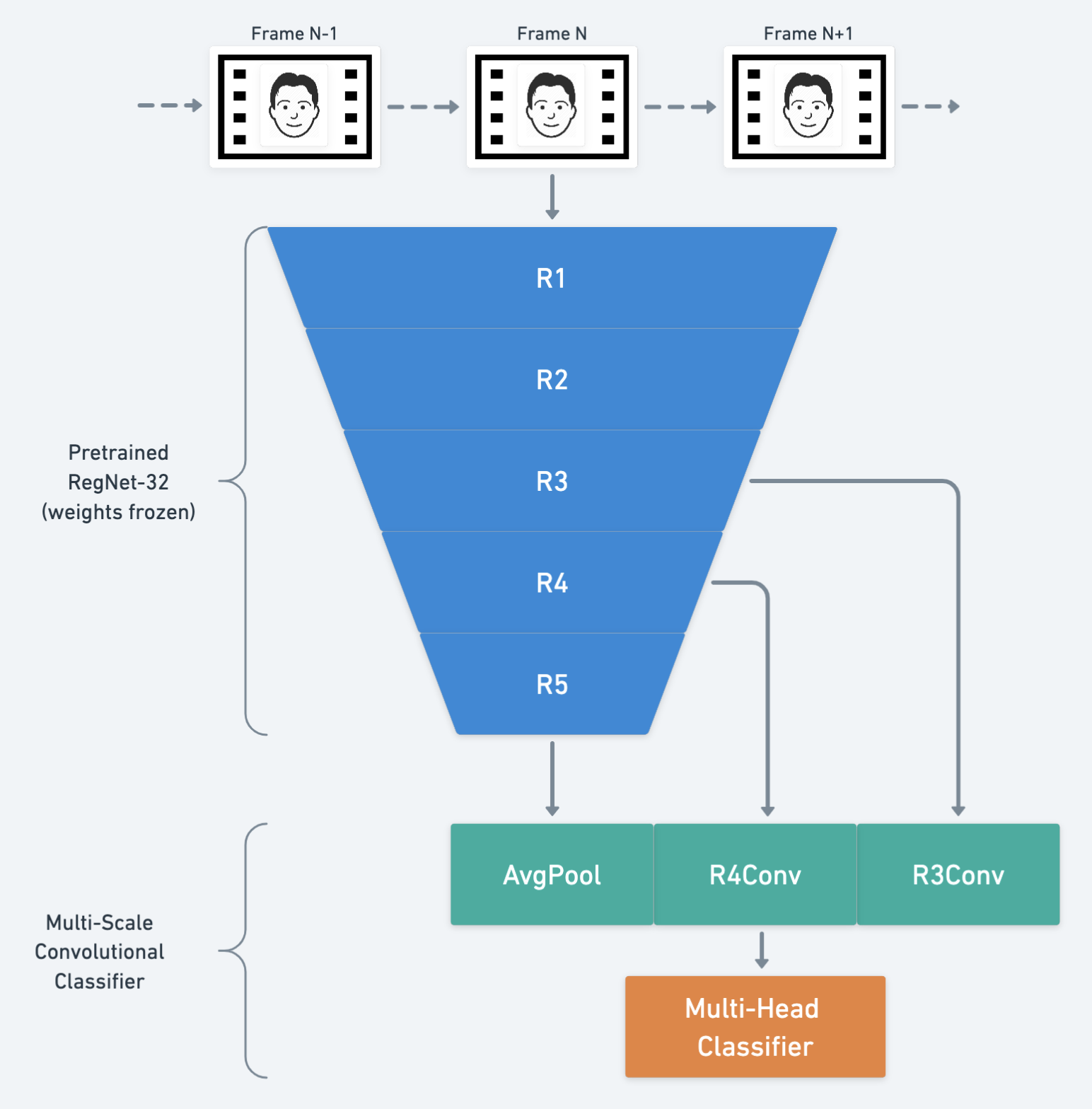}
   \caption{Network architecture. We apply the network for each frame in a video stream, sequentially before aggregating received predictions to get the final spoofing score for the entire session.}
   \label{fig:onecol}
\end{figure}

The training process is split into two steps in the following way:
\begin{enumerate}
    \item Task-aware fine-tuning (TAFT) of a larger network using Entry Antispoofing Dataset. 
    \item Distillation of a high-accuracy network into a smaller one, suitable for production use.
\end{enumerate}

\subsubsection{Task-aware fine-tuning}

Our training process follows the general approach described by Chen et al. in \cite{big-self-supervised}. To recap, their training pipeline starts with an unsupervised pretraining of a large network on a large amount of unlabeled data, followed by supervised fine-tuning on a (typically smaller) task-specific labeled dataset. In our experiments, an open source pretrained network provided by Meta \cite{SEER} substitutes the pretraining step. We do not fine-tune the layers imported from RegNet, following the established practice \cite{big-self-supervised} \cite{SEER} based on the observation that a large self-supervised pretrained network improves the generalization.

Our network structure is modified from RegNet-32g\cite{SEER}: for the final layer we use a multi-headed classifier composed from $8$ independent binary classifiers with the following semantics:

\begin{itemize}
    \item Heads 1-6 represent the explicitly defined, visible signals that the login session was being spoofed, i.e.: fingers holding a device, visible device border, mobile UI, moire patterns, screen glare, and reflections in the screen.
    \item Head 7 represents the probability of an attack that is not discernible to a human eye. In the dataset it meant that the video recording was a successful spoofing attempt, but no visible signals were identified.
    \item Head 8 represents the overall probability that the frame is fraudulent.
\end{itemize}

The training is done with the pretrained layers from \cite{SEER} frozen. We are optimizing the Reduced Focal Loss function \cite{ReducedFocalLoss} with AdamW\cite{AdamW} for 3 epochs with learning rate set to $1e-6$. Frames that are samples from source videos are heavily augmented to further prevent the domain shift with the following set of augmentations:
\begin{itemize}
    \item Random crop (ratio $0.33 - 1.0$ with random resize (ratio $0.7 - 1.35$).
    \item Random HSV shift
    \item Random Gaussian noise, motion blur
    \item Random ISO noise, 90 degree rotaion, horizontal flip
\end{itemize}

During training, we optimize a multi-class classification loss function for heads 1-8, but during inference only the probability from Head 8 is used. We observe that introducing an explicit classification of visible signals in a spoofing attempt improves the convergence speed and stabilizes the training process, but the actual spoofing detection does not require fine-grained classification. 

\subsubsection{Network distillation}

After the network is trained, we perform the distillation procedure with a smaller architecture, to make the real-time inference feasible. Since we have a large number of labeled videos in our training dataset $\mathcal{D}$, we are leveraging the weighted distillation loss from \cite{big-self-supervised}:

\begin{equation}
\begin{aligned}
    \mathcal{L} = & -(1-\alpha) \sum_{(x_i, y_i) \in \mathcal{D}^L} \Big[\log \mathit{P}^S(y_i|x_i)\Big] \\
    - & \alpha \sum_{x_i \in \mathcal{D}} \Big[ 
    \sum_y \mathit{P}^T(y|x_i;\tau)\log \mathit{P}^S(y|x_i;\tau)
    \Big]
\end{aligned}
\end{equation}
where $\tau$ is a scalar temperature parameter, $\alpha$ is a balancing parameter, $\mathit{P}^T(y|x_i)$ is the output of the teacher network, which is frozen after training, and $\mathit{P}^S(y|x_i)$ is the output of the student network.

The architecture for the student network is EfficientNet-B3\cite{efficientnet}.

\section{Experiments}

\subsection{Benchmarks}

We evaluate the effectiveness of our network from two aspects: an evaluation portion of Entry Antispoofing Dataset for intra-test, and MSU-MFSD, Replay-Attack, and OULU-NPU for cross-database tests. When it comes to the baselines, we compare the results of our models that were never fine-tuned on target datasets to both intra-test and cross-test results of existing state-of-the-art methods, indicating which is which in the tables. When the prior cross-test result is reported for a particular baseline, we choose the best score of all cross-tests for that model.

\paragraph{}\textbf{MSU-MFSD}. This dataset contains 280 videos of 35 subjects. Despite smaller scale, this benchmark is more challenging due to higher average quality of recordings used for spoofing, which is reflected in the baseline scored cited in this work.

\paragraph{} \textbf{Replay-Attack}. This dataset contains 1300 videos of 50 subjects. All videos are generated by either having a (real) client trying to access a laptop through a built-in webcam or by displaying a photo or a video recording of the same client for at least 9 seconds.

\paragraph{} \textbf{OULU-NPU}. This dataset contains 4950 videos of 55 subjects, collected in different lighting conditions, on six different mobile devices.

\subsection{Protocol}

The main goal of this work is to demonstrate strong generalization of our model across several challenging benchmarks. 

\subsection{Metric}

Our reported metric for cross-test evaluation is HTER (Half Total Error Rate)\cite{HTER}, which is widely used for comparing models in the field of biometric anti-spoofing. It is defined in terms of two error rates, False Acceptance Rate (FAR) and False Rejection Rate (FRR):

\begin{equation}
    HTER = \frac{FAR + FRR}{2}
\end{equation}

\section{Results}

%
%

In this paper we are reporting the results on two models, sharing the same general architecture and trained on the same data, with one architectural difference:

\begin{enumerate}
    \item \textbf{Entry V1}. This model predicts only the probability of spoofing directly, without utilizing explicitly defined features as described in section "Task-aware fine-tuning". Instead, this model is trained as a simple binary classification CNN.
    \item \textbf{Entry V2}. This model is trained according to the protocol from "Task-aware fine-tuning".
\end{enumerate}

\subsection{Intra-test on Entry Dataset}

First, we examine the results of the intra-test on a test subset of the internal Entry Antispoofing Dataset.

\begin{table}[h]
  \centering
  \begin{tabular}{@{}lc@{}}
    \toprule
    \textbf{Method} & \textbf{Intra-test on Entry} \\
    \midrule
    \midrule
    Entry-V1 & 3.54 \\
    \textbf{Entry-V2} & \textbf{0.74} \\
    \bottomrule
  \end{tabular}
  \caption{HTER(\%) scores on internal test subset of Entry Antispoofing Dataset. }
  \label{tab:intra-test}
\end{table}

Our primary hypothesis related to the dataset is that if the model achieving low HTER scores on it is capable of achieving similarly low error rates on other datasets, it will attest to the high level of generalization across domains it provides, and suggest that this dataset could be used on its own for comprehensive quality assessment moving forward.

\subsection{Cross-test}

We consider the generalizability of the model to be the main goal of building an accurate anti-spoofing algorithm, which is why we make the emphasis on cross-database testing. Following the established practice for conducting cross-database evaluation, we evaluate HTER scores on three challenging datasets: MSU-MFSD \cite{MFSD}, Replay-Attack \cite{mci/Chingovska2012}, OULU-NPU \cite{OULU_NPU_2017}. For the OULU-NPU evaluation, we chose Protocol I to be able to compare with the existing state-of-the-art results.

Model \textbf{Entry V2} achieves $HTER = 0$ on MSU-MFSD and Replay-Attack, therefore, it's possible to make a direct comparison with the results obtained on intra-tests that use Equal Error Rate metric, which is equivalen to $HTER$ at $EER = 0$.

\begin{table}[h]
  \centering
  \begin{tabular}{@{}lc@{}}
    \toprule
    \textbf{Method} & \textbf{MSU-MFSD} \\
    \midrule
    \midrule
    $CNN-LSTM^{AM}$ (Replay $\rightarrow$ MFSD)\cite{CNN-LSTM} & 25.72 \\
    GFA-CNN (Replay $\rightarrow$ MFSD)\cite{GFA-CNN} & 23.5 \\
    Entry-V1 & 2.4 \\
    \textbf{Entry-V2} & \textbf{0} \\
    \bottomrule
  \end{tabular}
  \caption{Cross-dataset HTER(\%) scores on MSU-MFSD\cite{MFSD}.}
  \label{tab:cross-test}
\end{table}

\begin{table}[h]
  \centering
  \begin{tabular}{@{}lc@{}}
    \toprule
    \textbf{Method} & \textbf{Replay-Attack} \\
    \midrule
    \midrule
    $CNN-LSTM^{AM}$ (MFSD $\rightarrow$ Replay)\cite{CNN-LSTM} & 12.37 \\
    GFA-CNN (CASIA $\rightarrow$ Replay)\cite{GFA-CNN} & 21.4 \\
    GFA-CNN (MFSD $\rightarrow$ Replay)\cite{GFA-CNN} & 25.8 \\
    CNCN (CASIA $\rightarrow$ Replay)\cite{CDCNN} & 15.5 \\
    CNCN++ (CASIA $\rightarrow$ Replay)\cite{CDCNN} & 6.5 \\
    Entry-V1 & 2.7 \\
    \textbf{Entry-V2} & \textbf{0} \\
    \bottomrule
  \end{tabular}
  \caption{Cross-dataset HTER(\%) scores on Replay-Attack\cite{mci/Chingovska2012}.}
  \label{tab:cross-test}
\end{table}

\begin{table}[h]
  \centering
  \begin{tabular}{@{}lc@{}}
    \toprule
    \textbf{Method} & \textbf{OULU-NPU} \\
    \midrule
    \midrule
    A-DeepPixBis (Replay $\rightarrow$ OULU)\cite{DeepPixBis} & 25.57 \\
    DeepPixBiS (Replay $\rightarrow$ OULU)\cite{DeepPix} & 22.7 \\
    Bi-FAS-S (Replay $\rightarrow$ OULU)\cite{Bi-FPNFAS} & 21.24 \\
    Bi-FAS (Replay $\rightarrow$ OULU)\cite{Bi-FPNFAS} & 18.33 \\
    LBP-SVM (Replay $\rightarrow$ OULU)\cite{DeepPix} & 12.1 \\
    IQM-SVM (Replay $\rightarrow$ OULU)\cite{IQM} & 3.9 \\
    \textbf{Entry-V1 (ours)} & \textbf{5.6} \\
    \textbf{Entry-V2 (ours} & \textbf{2.6} \\
    \bottomrule
  \end{tabular}
  \caption{Cross-dataset HTER(\%) scores comparison on Protocol I of the OULU-NPU dataset.}
  \label{tab:cross-test}
\end{table}

Additionally, for OULU-NPU we report the comparison on ACER metric.

\begin{table}[h]
  \centering
  \begin{tabular}{@{}lc@{}}
    \toprule
    \textbf{Method} & \textbf{OULU-NPU (ACER)} \\
    \midrule
    \midrule
    LBP-SVM (intra-test)\cite{DeepPix} & 25.0 \\
    IQM-SVM (intra-test)\cite{IQM} & 32.29 \\
    A-DeepPixBis (intra-test)\cite{DeepPixBis} & 0.75 \\
    \textbf{DeepPixBiS (intra-test)}\cite{DeepPix} & \textbf{0.42} \\
    Bi-FAS-S (intra-test)\cite{Bi-FPNFAS} & 1.97\\
    Bi-FAS (intra-test)\cite{Bi-FPNFAS} & 3.12\\
    Entry-V1 (ours) & 3.33 \\
    Entry-V2 (ours) & 3.2 \\
    \bottomrule
  \end{tabular}
  \caption{Cross-dataset ACER(\%) scores on Protocol I of the OULU-NPU dataset for more direct comparison with existing intra-tests. All models except ours were trained on OULU specifically. This comparison illustrates the closing gap between the results obtained by a strongly generalizable model (Entry) and the ones trained exclusively on OULU.}
  \label{tab:cross-test}
\end{table}

\subsection{Application}

This work was done as a part of R\&D effort inside XIX.ai supporting the key technology behind our biometric authentication system Entry. That was the reason why the generalization and performance requirements were dictated by the real-world applicability. 

\begin{enumerate}
    \item \textbf{High accuracy after distillation}. While the initial large network produces remarkable results, its size makes the scalability inefficient and, depending on the GPU accelerator used, prohibitively expensive. We have found that knowledge distillation does not have a noticeable effect on the model's accuracy when fine-tuned on Entry Antispoofing.
    \item \textbf{Real-time inference}. The anti-spoofing network, as all security-critical components, is being run on the backend, using cloud-hosted GPU accelerators. Since it turned out to be possible to distill the large model into a lightweight EfficientNet-B3, the throughput capacity was more than enough for processing multiple parallel video streams.
\end{enumerate}

\section{Conclusions}

In this paper we have presented an approach to training highly generalizable neural networks for face anti-spoofing, outlined the requirements for collecting labeled data sufficient for achieving the level of accuracy required for secure biometric authentication, and described the approach for post-processing of a model needed for deploying it as a part of a real-time video processing pipeline. We have shown the significant increase in accuracy on multiple established benchmarks by achieving the new state-of-the-art results from a combination of unsupervised pretraining and fine-tuning for the specific problem.

Augmenting the training objective with classification outputs predicting specific attributes of a spoofing attack alongside with the probability of an attack itself consistently improves the accuracy of the model further, without compromising on the generalization.

Finally, we have tested the performance differences and the increase in the inference throughput after the model distillation to prove the viability of this model in a live application.

{\small
\bibliographystyle{ieee_fullname}
\bibliography{paper}
}

\end{document}